%% file: main.tex
\documentclass[sigconf]{acmart}

\usepackage{booktabs} 
\usepackage{graphicx}
\usepackage{hyperref}
\usepackage{booktabs}
\usepackage{amssymb}
\usepackage{tablefootnote}
\usepackage{mdframed}
\usepackage{enumitem}
\usepackage{tabu}
\usepackage{subcaption}
\usepackage{multirow}
\usepackage{xcolor}

\usepackage[skip=0.5\baselineskip]{caption}

\usepackage{cleveref}
\crefformat{chapter}{\S#2#1#3}
\crefmultiformat{chapter}{\S\S#2#1#3}{and~#2#1#3}{, #2#1#3}{, and~#2#1#3}

\crefformat{section}{\S#2#1#3}
\crefmultiformat{section}{\S\S#2#1#3}{, #2#1#3}{, #2#1#3}{, and~#2#1#3}

\newcommand{\sep}{\textcolor{gray}{| }}
\newcommand{\semantictype}[1]{\texttt{#1}}

\usepackage{xspace}
\newcommand{\ie}{{i.e.,}\xspace}
\newcommand{\eg}{{e.g.,}\xspace}

\copyrightyear{2019} 
\acmYear{2019} 
\setcopyright{acmlicensed}
\acmConference[KDD '19]{The 25th ACM SIGKDD Conference on Knowledge Discovery and Data Mining}{August 4--8, 2019}{Anchorage, AK, USA}
\acmBooktitle{The 25th ACM SIGKDD Conference on Knowledge Discovery and Data Mining (KDD '19), August 4--8, 2019, Anchorage, AK, USA}
\acmPrice{15.00}
\acmDOI{10.1145/3292500.3330993}
\acmISBN{978-1-4503-6201-6/19/08}

\begin{document}
\title[Sherlock: A Deep Learning Approach to Semantic Data Type Detection]{Sherlock: A Deep Learning Approach to \\ Semantic Data Type Detection}

\author{Madelon Hulsebos}
\affiliation{%
  \institution{MIT Media Lab}
}
\email{madelonhulsebos@gmail.com}

\settopmatter{authorsperrow=4}
\author{Kevin Hu}
\affiliation{
 \institution{MIT Media Lab}
}
\email{kzh@mit.edu}

\author{Michiel Bakker}
\affiliation{%
  \institution{MIT Media Lab}
}
\email{bakker@mit.edu}

\author{Emanuel Zgraggen}
\affiliation{%
 \institution{MIT CSAIL}
}
\email{emzg@mit.edu}

\author{Arvind Satyanarayan}
\affiliation{%
 \institution{MIT CSAIL}
}
\email{arvindsatya@mit.edu}
 
\author{Tim Kraska}
\affiliation{%
  \institution{MIT CSAIL}
}
\email{kraska@mit.edu}

\author{{\c{C}}a{\u{g}}atay Demiralp}
\affiliation{%
 \institution{Megagon Labs}
}
\email{cagatay@megagon.ai}

\author{C{\'{e}}sar Hidalgo}
\affiliation{%
 \institution{MIT Media Lab}
}
\email{hidalgo@mit.edu}

\renewcommand{\shortauthors}{Hulsebos et al.}

\begin{abstract}
Correctly detecting the semantic type of data columns is crucial for data science tasks such as automated data cleaning, schema matching, and data discovery. Existing data preparation and analysis systems rely on dictionary lookups and regular expression matching to detect semantic types. However, these matching-based approaches often are not robust to dirty data and only detect a limited number of types. We introduce Sherlock, a multi-input deep neural network for detecting semantic types. We train Sherlock on $686,765$ data columns retrieved from the VizNet corpus by matching $78$ semantic types from DBpedia to column headers. We characterize each matched column with $1,588$ features describing the statistical properties, character distributions, word embeddings, and paragraph vectors of column values. Sherlock achieves a support-weighted F$_1$ score of $0.89$, exceeding that of machine learning baselines, dictionary and regular expression benchmarks, and the consensus of crowdsourced annotations. 
\end{abstract}



\begin{teaserfigure}
    \centering
    \includegraphics[width=\textwidth]{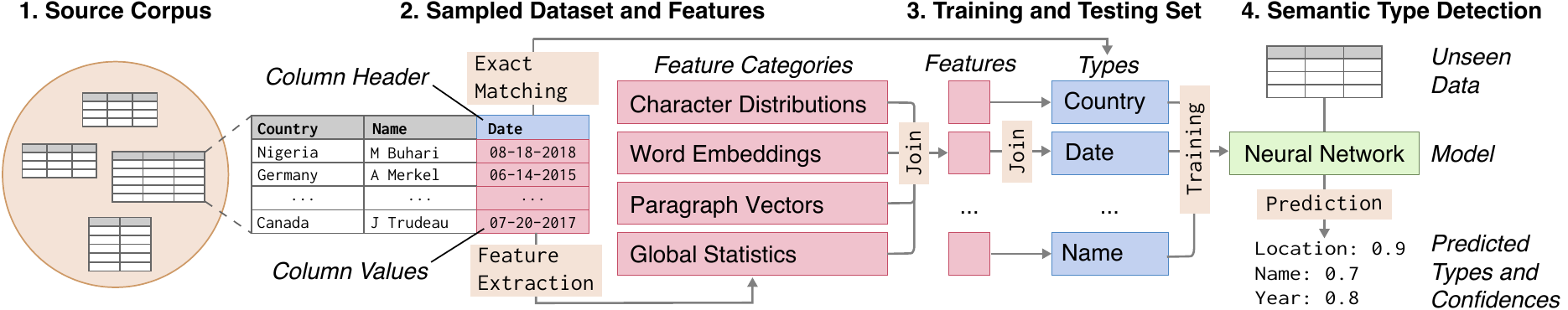}
    \caption{Data processing and analysis flow, starting from (1) a corpus of real-world datasets, proceeding to (2) feature extraction, (3) mapping extracted features to ground truth semantic types, and (4) model training and prediction.}
\end{teaserfigure}

\maketitle

\input{1_introduction.tex}
\input{2_related_work.tex}
\input{3_data_representation.tex}
\input{4_model.tex}
\input{5_results.tex}
\input{6_discussion.tex}
\input{8_conclusion.tex}

\bibliographystyle{ACM-Reference-Format}
\bibliography{bibliography}

\begin{appendix}
\input{A_appendix.tex}
\end{appendix}

\end{document}

%% file: 1_introduction.tex
\section{Introduction}

Data preparation and analysis systems rely on correctly detecting types of data columns to enable and constrain functionality. For example, automated data cleaning facilitates the generation of clean data through validation and transformation rules that depend on data type~\cite{2011-wrangler, Raman:2001:PWI:645927.672045}. Schema matching identifies correspondences between data objects, and frequently uses data types to constrain the search space of correspondences~\cite{Rahm:2001:SAA:767149.767154, zapilko}. Data discovery surfaces data relevant to a given query, often relying on semantic similarities across tables and columns~\cite{aurum, seeping-semantics}.

While most systems reliably detect \textit{atomic types} such as \texttt{string}, \texttt{integer}, and \texttt{boolean}, \textit{semantic types} are disproportionately more powerful and in many cases essential. Semantic types provide finer-grained descriptions of the data by establishing correspondences between columns and real-world concepts and as such, can help with schema matching to determine which columns refer to the same real-world concepts, or data cleaning by determining the conceptual domain of a column. 
In some cases, the detection of a semantic type can be easy. 
For example, an \semantictype{ISBN} or \semantictype{credit card number} are generated according to strict validation rules, lending themselves to straightforward type detection with just a few rules. 
But most types, including \semantictype{location}, \semantictype{birth date}, and \semantictype{name}, do not adhere to such structure, as shown in Table~\ref{tab:data_values_examples}.

\begin{table}[t]
    \caption{Data values sampled from real-world datasets.} 
    \label{tab:data_values_examples} 
    \begin{tabu}{lX}
        \toprule
        \textbf{Type} & \textbf{Sampled values} \\ \midrule
        \semantictype{location} & TBA \sep Chicago, Ill. \sep Detroit, Mich. \sep Nashville, Tenn. \\
        \semantictype{location} & UNIVERSITY SUITES \sep U.S. 27; N\/A \sep NORSE HALL\\
        \semantictype{location} & Away \sep  Away \sep  Home \sep  Away \sep  Away \\
        \semantictype{date} & 27 Dec 1811 \sep  1852 \sep  1855 \sep  - \sep  1848 \sep  1871 \sep  1877 \\
        \semantictype{date} & -- --, 1922 \sep  -- --, 1902 \sep  -- --, 1913 \sep  -- --, 1919 \\
        \semantictype{date} & December 06 \sep  August 23 \sep  None \\
        \semantictype{name} & Svenack \sep Svendd \sep  Sveneldritch \sep Sveng{\"o}ran \\
        \semantictype{name} & HOUSE, BRIAN \sep HSIAO, AMY \sep  HSU, ASTRID \\
        \semantictype{name} & D. Korb \sep  K. Moring \sep  J. Albanese \sep  l. dunn \\ \bottomrule
    \end{tabu}
\end{table}

\begin{figure}[h]
    \centering
    \includegraphics[width=\columnwidth]{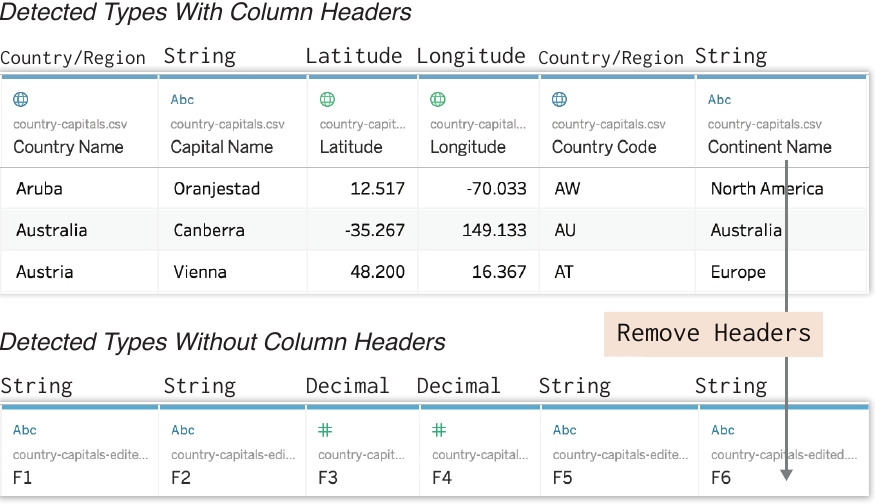}
    \caption{Data types detected by Tableau Desktop 2018.3 for a dataset of country capitals, with and without headers.}
    \label{fig:tableau}
\end{figure}

Existing open source and commercial systems take \textit{matching-based} approaches to semantic type detection. For example, regular expression matching captures patterns of data values using predefined character sequences. Dictionary approaches use matches between data headers and values with internal look-up tables. While sufficient for detecting simple types, these matching-based approaches are often not robust to malformed or dirty data, support only a limited number of types, and under-perform for types without strict validations. For example, Figure~\ref{fig:tableau} shows that Tableau detects a column labeled ``Continent Name'' as \semantictype{string}. After removing column headers, no semantic types are detected.
Note that missing headers or incomprehensible headers are not uncommon. 
For example, SAP's system table $T005$ contains country information and column $NMFMT$ is the standard name field, whereas $INTCA$ refers to the ISO code or $XPLZS$ to zip-code. 

Machine learning models, coupled with large-scale training and benchmarking corpora, have proven effective at predictive tasks across domains. Examples include the AlexNet neural network trained on ImageNet for visual recognition and the Google Neural Machine Translation system pre-trained on WMT parallel corpora for language translation. Inspired by these advances, we introduce \textbf{Sherlock}, a deep learning approach to semantic type detection trained on a large corpus of real-world columns.

To begin, we consider 78 semantic types described by T2Dv2 Gold Standard,\footnote{\url{http://webdatacommons.org/webtables/goldstandardV2.html}} which matches properties from the DBpedia ontology with column headers from the WebTables corpus. Then, we use exact matching between semantic types and column headers to extract $686,765$ data columns from the VizNet corpus \cite{hu2019viznet}, a large-scale repository of real world datasets collected from the web, popular visualization systems, and open data portals.

We consider each column as a mapping from column values to a column header. We then extract $1,588$ features from each column, describing the distribution of characters, semantic content of words and columns, and global statistics such as cardinality and uniqueness. Treating column headers as ground truth labels of the semantic type, we formulate semantic type detection as a multiclass classification problem.

A multi-input neural network architecture achieves a support-weighted F\textsubscript{1}-score of $0.89$, exceeding that of decision tree and random forest baseline models, two matching-based approaches that represent type detection approaches in practice, and the consensus of crowdsourced annotations. We then examine types for which the neural network demonstrates high and low performance, investigate the contribution of each feature category to model performance, extract feature importances from the decision tree baseline, and present an error-reject curve suggesting the potential of combining learned models with human annotations.

To conclude, we discuss promising avenues for future research in semantic type detection, such as assessing training data quality at scale, enriching feature extraction processes, and establishing shared benchmarks. To support benchmarks for future research and integration into existing systems, we open source our data, code, and trained model at \url{https://sherlock.media.mit.edu}. 

\vspace{0.25cm}
\begin{mdframed}[backgroundcolor=black!10]
\textbf{Key contributions}:
\begin{enumerate}[noitemsep, leftmargin=*, topsep=3pt]
    \item \textbf{Data} (\cref{sec:data}): Demonstrating a scalable process for matching $686,675$ columns from VizNet corpus for $78$ semantic types, then describing with $1,588$ features like word- and paragraph embeddings.

    \item \textbf{Model} (\cref{sec:model}): Formulating type detection as a multiclass classification problem, then contributing a novel multi-input neural network architecture.
    
    \item \textbf{Results} (\cref{sec:results}): Benchmarking predictive performance against a decision tree and random forest baseline, two matching-based models, and crowdsourced consensus. 
\end{enumerate}
\end{mdframed}

%% file: 2_related_work.tex
\section{Related Work}
\label{sec:related_work}

Sherlock is informed by existing \textit{commercial and open source systems} for data preparation and analysis, as well as prior research work on \textit{ontology-based}, \textit{feature-based}, \textit{probabilistic}, and \textit{synthesized} approaches to semantic type detection.

\paragraph{Commercial and open source}
Semantic type detection enhances the functionality of commercial data preparation and analysis systems such as Microsoft Power BI \cite{powerbi}, Trifacta \cite{trifacta}, and Google Data Studio \cite{googledatastudio}.
To the best of our knowledge, these commercial tools rely on manually defined regular expression patterns dictionary lookups of column headers and values to detect a limited set of semantic types. For instance, Trifacta detects around $10$ types (\eg \semantictype{gender} and \semantictype{zip code}) and Power BI only supports time-related semantic types (\eg \semantictype{date/time} and \semantictype{duration}). Open source libraries such as messytables \cite{messytables}, datalib \cite{datalib}, and csvkit \cite{csvkit} similarly use heuristics to detect a limited set of types. Benchmarking directly against these systems was infeasible due to the small number of supported types and lack of extensibility. However, we compare against learned regular expression and dictionary-based benchmarks representative of the approaches taken by these systems.

\paragraph{Ontology-based}
Prior research work, with roots in the semantic web and schema matching literature, provide alternative approaches to semantic type detection. One body of work leverages existing data on the web, such as WebTables \cite{webtables}, and ontologies (or, knowledge bases) such as DBPedia \cite{dbpedia}, Wikitology \cite{syed2010exploiting}, and Freebase~\cite{freebase}. Venetis et al. \cite{venetis2011recovering} construct a database of value-type mappings, then assign types using a maximum likelihood estimator based on column values. Syed et al. \cite{syed2010exploiting} use column headers and values to build a Wikitology query, the result of which maps columns to types. Informed by these approaches, we looked towards existing ontologies to derive the 275 semantic types considered in this paper.

\paragraph{Feature-based}
Several approaches capture and compare properties of data in a way that is ontology-agnostic. Ramnandan et al.~\cite{ramnandan2015assigning} use heuristics to first separate numerical and textual types, then describe those types using the Kolmogorov-Smirnov (K-S) test and Term Frequency-Inverse Document Frequency (TF-IDF), respectively. Pham et al.~\cite{pham2016semantic} use slightly more features, including the Mann-Whitney test for numerical data and Jaccard similarity for textual data, to train logistic regression and random forest models. We extend these feature-based approaches with a significantly larger set of features that includes character-level distributions, word embeddings, and paragraph vectors. We leverage orders of magnitude more features and training samples than prior work in order to train a high-capacity machine learning model, a deep neural network. 
We include a decision tree and random forest model as benchmarks to represent these ``simpler'' machine learning models.

\paragraph{Probabilistic}

The third category of prior work employs a probabilistic approach. Goel et al. \cite{goel2012exploiting} use conditional random fields to predict the semantic type of each value within a column, then combine these predictions into a prediction for the whole column. Limaye et al. \cite{limaye2010annotating} use probabilistic graphical models to annotate values with entities, columns with types, and column pairs with relationships. These predictions simultaneously maximize a potential function using a message passing algorithm. Probabilistic approaches are complementary to our machine learning-based approach by providing a means for combining column-specific predictions. However, as with prior feature-based models, code for retraining these models was not made available for benchmarking.

\paragraph{Synthesized}
Puranik~\cite{puranik} proposes a ``specialist approach'' combining the predictions of regular expressions, dictionaries, and machine learning models. More recently, Yan and He \cite{yan2018synthesizing} introduced a system that, given a search keyword and set of positive examples, synthesizes type detection logic from open source GitHub repositories. This system provides a novel approach to leveraging domain-specific heuristics for parsing, validating, and transforming semantic data types. While both approaches are exciting, the code underlying these systems was not available for benchmarking.

%% file: 3_data_representation.tex
\section{Data}\label{sec:data}

We describe the semantic types we consider, how we extracted data columns from a large repository of real-world datasets, and our feature extraction procedure.

\subsection{Data Collection}

Ontologies like WordNet~\cite{wordnet} and DBpedia~\cite{dbpedia} describe semantic concepts, properties of such concepts, and relationships between them. To constrain the number of types we consider, we adopt the types described by the T2Dv2 Gold Standard,\footnotemark[1] the result of a study matching DBpedia properties~\cite{ritze2017matching} with columns from the Web Tables web crawl corpus~\cite{webtables}. These $275$ DBpedia properties, such as \semantictype{country}, \semantictype{language}, and \semantictype{industry}, represent semantic types commonly found in datasets scattered throughout the web.

To expedite the collection of real-world data from diverse sources, we use the VizNet repository~\cite{hu2019viznet}, which aggregates and characterizes data from two popular online visualization platforms and open data portals, in addition to the Web Tables corpus. For feasibility, we restricted ourselves to the first $10$M Web Tables datasets, but considered the remainder of the repository in its entirety. We then match data columns from VizNet that have headers corresponding to our 275 types. To accomodate variation in casing and formatting, single word types matched case-altered modifications (\eg \semantictype{name} = \semantictype{Name} = \semantictype{NAME}) and multi-word types included concatenations of constituent words (\eg \semantictype{release date} = \semantictype{releaseDate}).

The matching process resulted in $6,146,940$ columns matching the $275$ considered types. Manual verification indicated that the majority of columns were plausibly described by the corresponding semantic type, as shown in Table~\ref{tab:data_values_examples}. In other words, matching column headers as ground truth labels of the semantic type yielded high quality training data.

\subsection{Feature Extraction}
\label{sec:features}

To create fixed-length representations of variable-length columns, aid interpretation of results, and provide ``hints'' to our neural network, we extract features from each column. To capture different properties of columns, we extract four categories of features: global statistics (27), aggregated character distributions (960), pretrained word embeddings (200), and self-trained paragraph vectors (400).

\paragraph{Global statistics}\label{sec:stat_features}
The first category of features describes high-level statistical characteristics of columns. For example, the ``column entropy'' feature describes how uniformly values are distributed. Such a feature helps differentiate between types that contain more repeated values, such as \semantictype{gender}, from types that contain many unique values, such as \semantictype{name}. Other types, like \semantictype{weight} and \semantictype{sales}, may consist of many numerical characters, which is captured by the ``mean of the number of numerical characters in values.'' A complete list of these $27$ features can be found in Table~\ref{tab:character-features} in the Appendix.

\paragraph{Character-level distributions}
Preliminary analysis indicated that simple statistical features such as the ``fraction of values with numerical characters'' provide surprising predictive power. Motivated by these results and the prevalence of character-based matching approaches such as regular expressions, we extract features describing the distribution of characters in a column. Specifically, we compute the count of all $96$ ASCII-printable characters (\ie digits, letters, and punctuation characters, but not whitespace) within each value of a column. We then aggregate these counts with 10 statistical functions (\ie any, all, mean, variance, min, max, median, sum, kurtosis, skewness), resulting in $960$ features. Example features include ``whether all values contain a `-' character'' and the ``mean number of `/' characters.''

\paragraph{Word embeddings}
For certain semantic types, columns frequently contain commonly occurring words. For example, the \semantictype{city} type contains values such as \textit{New York City}, \textit{Paris}, and \textit{London}. To characterize the semantic content of these values, we used word embeddings that map words to high-dimensional fixed-length numeric vectors. In particular, we used a pre-trained GloVe dictionary~\cite{pennington2014glove} containing $50$-dimensional representations of $400$K English words aggregated from $6$B tokens, used for tasks such as text similarity~\cite{kenter2015short}. For each value in a column, if the value is a single word, we look up the word embedding from the GloVe dictionary. We omit a term if it does not appear in the GloVe dictionary. For values containing multiple words, we looked up each distinct word and represented the value with the mean of the distinct word vectors. Then, we computed the mean, mode, median and variance of word vectors across all values in a column.

\paragraph{Paragraph vectors}
To represent each column with a fixed-length numerical vector, we implemented the Distributed Bag of Words version of Paragraph Vector (PV-DBOW)~\cite{le2014distributed}. Paragraph vectors were originally developed to numerically represent the ``topic'' of pieces of texts, but have proven effective for more general tasks, such as document similarity~\cite{dai2015document}. In our implementation, each column is a ``paragraph'' while values within a column are ``words'': both the entire column and constituent values are represented by one-hot encoded vectors.

After pooling together all columns across all classes, the training procedure for each column in the same 60\% training set used by the main Sherlock model is as follows. We randomly select a window of value vectors, concatenate the column vector with the remaining value vectors, then train a single model to predict the former from the latter. Using the Gensim library~\cite{gensim2010}, we trained this model for 20 iterations. We used the trained model to map each column in both the training and test sets to a $400$-dimensional paragraph vector, which provided a balance between predictive power and computational tractability.

\subsection{Filtering and Preprocessing}
Certain types occur more frequently in the VizNet corpus than others. For example, \semantictype{description} and \semantictype{city} are more common than \semantictype{collection} and \semantictype{continent}. To address this heterogeneity, we limited the number of columns to at most $15$K per class and excluded the $10\%$ types containing less than $1$K columns.

\begin{figure}[h]
    \centering
    \hspace*{-0.1cm}
    \includegraphics[width=\columnwidth]{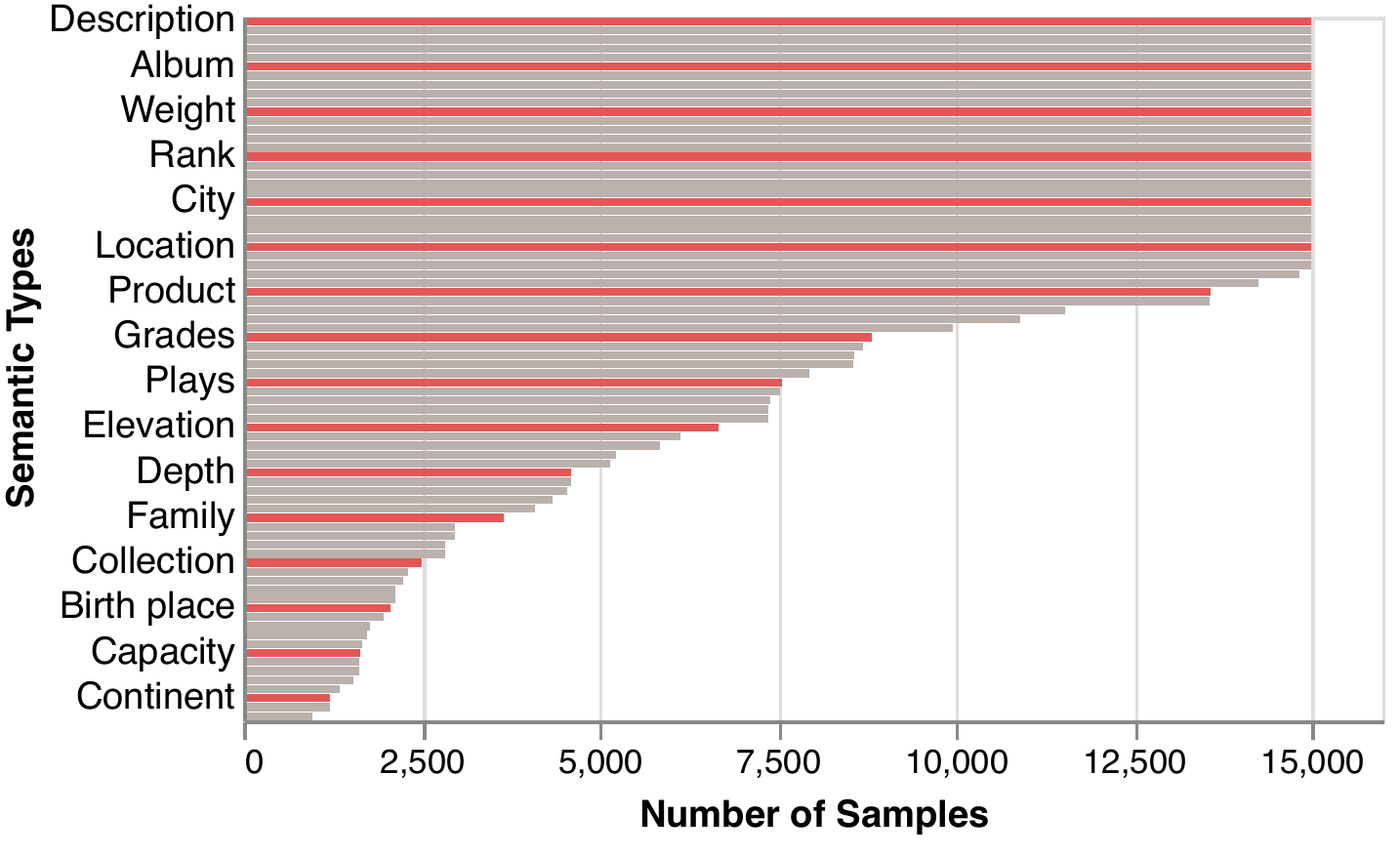}
    \caption{
    Number of columns per semantic type extracted from VizNet after filtering out the types with more than 15\% of the columns not present in the GloVe dictionary, or with less than 1K columns.
    }
    \label{fig:distribution_classes}
\end{figure}

Other semantic types, especially those describing numerical concepts, are unlikely to be represented by word embeddings. To contend with this issue, we filtered out the types for which at least $15\%$ of the columns did not contain a single word that is present in the GloVe dictionary. This filter resulted in a final total of \textbf{686,765 columns} corresponding to \textbf{78 semantic types}, of which a list is included in Table~\ref{tab:classes} in the Appendix. The distribution of number of columns per semantic type is shown in Figure~\ref{fig:distribution_classes}. 

Before modeling, we preprocess our features by creating an additional binary feature indicating whether word embeddings were successfully extracted for a given column. Including this feature results in a total of \textbf{1,588 features}. Then, we impute missing values across all features with the mean of the respective feature.

%% file: 4_model.tex
\section{Methods}\label{sec:model}

We describe our deep learning model, random forest baseline, two matching-based benchmarks, and crowdsourced consensus benchmark. Then, we explain our training and evaluation procedures.

\subsection{Sherlock: A Multi-input Neural Network}

\begin{figure}[h]
    \includegraphics[width=\columnwidth]{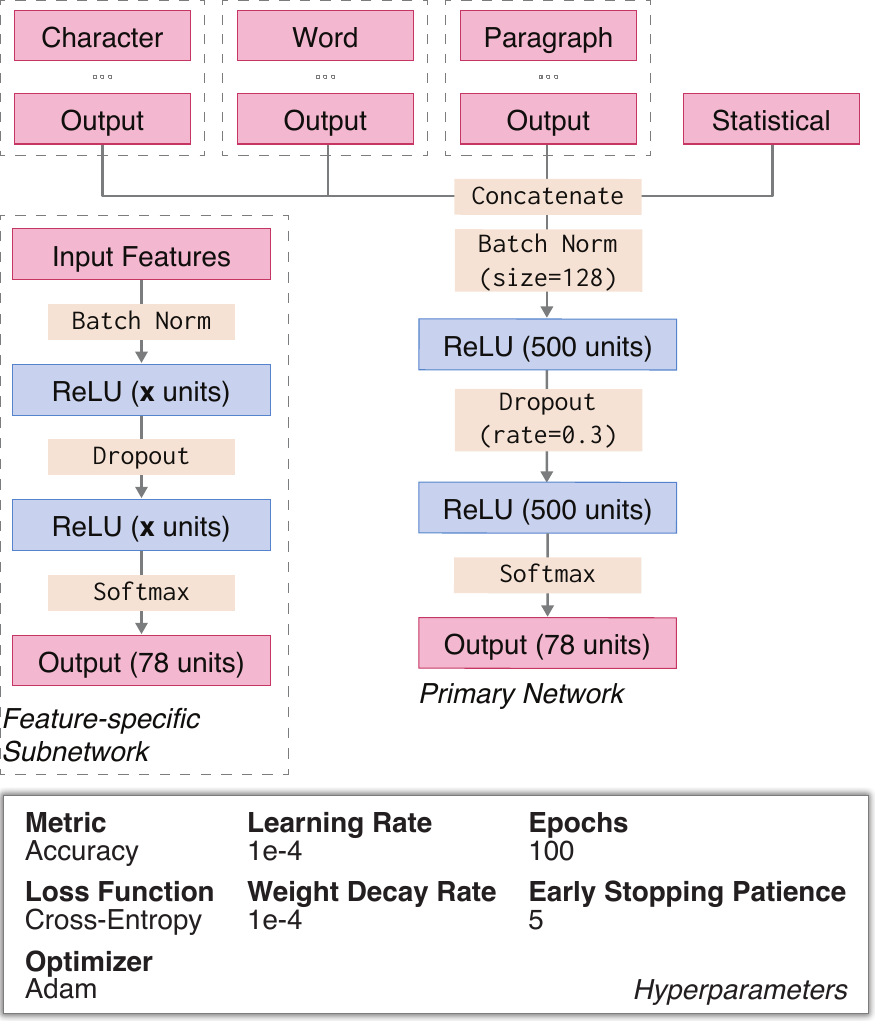}
    \caption{
    Architecture of the primary network and its feature-specific subnetworks, and the hyperparameters used for training. 
    }
    \label{fig:architecture}
\end{figure}

Prior machine learning approaches to semantic type detection \cite{limaye2010annotating, venetis2011recovering} trained simple models, such as logistic regression, on relatively small feature sets. We consider a significantly larger number of features and samples, which motivates our use of a feedforward neural network. Specifically, given the different number of features and varying noise levels within each feature category, we use a multi-input architecture with hyperparameters shown in Figure~\ref{fig:architecture}.

At a high-level, we train subnetworks for each feature category except the statistical features, which consist of only $27$ features. These subnetworks ``compress'' input features to an output of fixed dimension. We chose this dimension to be equal to the number of types in order to evaluate each subnetwork independently. Then, we concatenate the weights of the three output layers with the statistical features to form the input layer of the primary network.

Each network consists of two hidden layers with rectified linear unit (ReLU) activation functions. Experiments with hidden layer sizes between $100$ and $1,000$ (\ie on the order of the input layer dimension) indicate that hidden layer sizes of $300$, $200$, and $400$ for the character-level, word embedding, and paragraph vector subnetworks, respectively, provides the best results. To prevent overfitting, we included drop out layers and weight decay terms. The final class predictions result from the output of the final softmax layer, corresponding to the network's confidence about a sample belonging to each class, the predicted label then is the class with the highest confidence. The neural network, which we refer to as ``Sherlock,'' is implemented in TensorFlow~\cite{tensorflow2015-whitepaper}.

\subsection{Benchmarks}\label{sec:benchmarks}
To measure the relative performance of Sherlock, we compare against four benchmarks.

\paragraph{Machine learning classifiers}\label{sec:baseline}
The first benchmark is a decision tree, a non-parametric machine learning model with reasonable ``out-of-the-box'' performance and straightforward interpretation. We use the decision tree to represent the simpler models found in prior research, such as the logistic regression used in Pham et al.~\cite{pham2016semantic}. Learning curves indicated that decision tree performance plateaued beyond a depth of $50$, which we then used as the maximum depth. We also add a random forest classifier we built from $10$ such trees, which often yields significantly better performance. For all remaining parameters, we used the default settings in the scikit-learn package~\cite{scikit-learn}.

\begin{figure}[h]
    \includegraphics[width=\columnwidth]{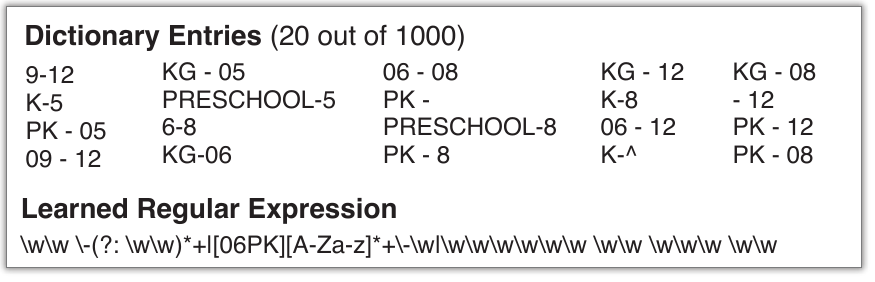}
    \caption{Examples of dictionary entries and a learned regular expression for the \semantictype{grades} type.}
    \label{fig:example_dict_regex}
\end{figure}

\paragraph{Dictionary}
Dictionaries are commonly used to detect semantic types that contain a finite set of valid values, such as \semantictype{country}, \semantictype{day}, and \semantictype{language}. The first matching-based benchmark is a dictionary that maps column values or headers to semantic types. For each type, we collected the $1,000$ most frequently occurring values across all columns, resulting in $78,000$ \{ \semantictype{value} : \semantictype{type} \} pairs. For example, Figure~\ref{fig:example_dict_regex} shows examples of entries mapped to the \semantictype{grades} type. Given an unseen data column at test time, we compare $1,000$ randomly selected column values to each entry of the dictionary, then classify the column as the most frequently matched type. 

\paragraph{Learned regular expressions}
Regular expressions are frequently used to detect semantic types with common character patterns, such as \semantictype{address}, \semantictype{birth date}, and \semantictype{year}. The second matching-based benchmark uses patterns of characters specified by learned regular expressions. We learn regular expressions for each type using the evolutionary procedure of Bartoli et al.~\cite{bartoli2016inference}. Consistent with the original setup, we randomly sampled $50$ ``positive values'' from each type, and $50$ ``negative'' values from other types. An example of a learned regular expression in Java format for the \semantictype{grades} type is shown in Figure~\ref{fig:example_dict_regex}. As with the dictionary benchmark, we match $1,000$ randomly selected values against learned regular expressions, then use majority vote to determine the final predicted type.

\paragraph{Crowdsourced annotations}
To assess the performance of human annotators at predicting semantic type, we conducted a crowdsourced experiment. The experiment began by defining the concepts of data and semantic type, then screened out participants unable to select a specified semantic type. After the prescreen, participants completed three sets of ten questions separated by two attention checks. Each question presented a list of data values, asked ``Which one of the following types best describes these data values?'', and required participants to select a single type from a scrolling menu with 78 types. Questions were populated from a pool of $780$ samples containing $10$ randomly selected values from all $78$ types.

We used the Mechanical Turk crowdsourcing platform~\cite{mturk} to recruit $390$ participants that were native English speakers and had ${\ge}95\%$ HIT approval rating, ensuring high-quality annotations. Participants completed the experiment in 16 minutes and 22 seconds on average and were compensated 2 USD, a rate slightly exceeding the United States federal minimum wage of 7.25 USD. Detailed worker demographics are described in Appendix~\ref{app:mturk_demographics}. Overall, $390$ participants annotated $30$ samples each, resulting in a total of $11,700$ annotations, or an average of $15$ annotations per sample. For each sample, we used the most frequent (\ie the mode) type from the $15$ annotations as the crowdsourced consensus annotation.

\subsection{Training and Evaluation}
To ensure consistent evaluation across benchmarks, we divided the data into 60/20/20 training/validation/testing splits. To account for class imbalances, we evaluate model performance using the average $\text{F$_1$-score} = 2 \times (\text{precision} \times \text{recall}) / (\text{precision} + \text{recall})$, weighted by the number of columns per class in the test set (\ie the support). To estimate the mean and $95\%$ percentile error of the crowdsourced consensus F$_1$ score, we conducted $10^5$ bootstrap simulations by resampling annotations for each sample with replacement.

Computational effort and space required at prediction time are also important metrics for models incorporated into user-facing systems. We measure the average time in seconds needed to extract features and generate a prediction for a single sample, and report the space required by the models in megabytes.

%% file: 5_results.tex
\section{Results}\label{sec:results}

We report the performance of our multi-input neural network and compare against benchmarks. Then, we examine types for which Sherlock demonstrated high and low performance, the contribution of each feature category in isolation, decision tree feature importances, and the effect of rejection threshold on performance.

\subsection{Benchmark Results}
We compare Sherlock against decision tree, random forest, dictionary-based, learned regular expression, and crowdsourced consensus benchmarks. Table~\ref{tab:performances_benchmarks} presents the F$_1$ score weighted by support, runtime in seconds per sample, and size in megabytes of each model.

\begin{table}[h]
    \centering
    \caption{Support-weighted F$_1$ score, runtime at prediction, and size of Sherlock and four benchmarks.}
    \label{tab:performances_benchmarks}    
    \begin{tabu}{ X r r r }
        \toprule
        \textbf{Method} & \textbf{F$_1$ Score} & \textbf{Runtime (s)} & \textbf{Size (Mb)} \\ \hline \addlinespace[0.1em]
        \multicolumn{4}{c}{\textit{Machine Learning}} \\ \midrule
        Sherlock & $0.89$ & $0.42$ ($\pm 0.01$) & $6.2$ \\
        Decision tree & $0.76$ & $0.26$ ($\pm 0.01$) & $59.1$ \\
        Random forest & $0.84$ & $0.26$ ($\pm 0.01$) & $760.4$ \\
        \midrule
        \multicolumn{4}{c}{\textit{Matching-based}} \\\midrule
        Dictionary & $0.16$ & $0.01$ ($\pm 0.03$) & $0.5$ \\
        Regular expression & $0.04$  & $0.01$ ($\pm 0.03$) & $0.01$ \\ \hline \addlinespace[0.1em]
        \multicolumn{4}{c}{\textit{Crowdsourced Annotations}} \\ \midrule
        Consensus & $0.32$ ($\pm 0.02$) & $33.74$ ($\pm 0.86$) & $-$ \\ 
        \bottomrule
    \end{tabu}
\end{table}

We first note that the machine learning models significantly outperform the matching-based and crowdsourced consensus benchmarks, in terms of F$_1$ score. The relatively low performance of crowdsourced consensus is perhaps due to the visual overload of selecting from 78 types, such that performance may increase with a smaller number of candidate types. Handling a large number of candidate classes is a benefit of using an ML-based or matching-based model. Alternatively, crowdsourced workers may have difficulties differentiating between classes that are unfamiliar or contain many numeric values. Lastly, despite our implementing basic training and honeypot questions, crowdsourced workers will likely improve with longer training times and stricter quality control.

Inspection of the matching-based benchmarks suggests that dictionaries and learned regular expressions are prone to ``overfitting'' on the training set. Feedback from crowdsourced workers suggests that annotating semantic types with a large number of types is a challenging and ambiguous task.

Comparing the machine learning models, Sherlock significantly outperforms the decision tree baseline, while the random forest classifier is competitive. 
For cases in which interpretability of features and predictions are important considerations, the tree-based benchmarks may be a suitable choice of model.

Despite poor predictive performance, matching-based benchmarks are significantly smaller and faster than both machine learning models. For cases in which absolute runtime and model size are critical, optimizing matching-based models may be a worthwhile approach. This trade-off also suggests a hybrid approach of combining matching-based models for ``easy'' types with machine learning models for more ambiguous types.

\subsection{Performance for Individual Types}

Table~\ref{tab:top_five_and_bottom_five_types} displays the top and bottom five types, as measured by the F$_1$ score achieved by Sherlock for that type. High performing types such as \semantictype{grades} and \semantictype{industry} frequently contain a finite set of valid values, as shown in Figure~\ref{fig:example_dict_regex} for \semantictype{grades}. Other types such as \semantictype{birth date} and \semantictype{ISBN}, often follow consistent character patterns, as shown in Table~\ref{tab:data_values_examples}. 

\begin{table}[h]
\caption{Top five and bottom five types by F$_1$ score.}
\label{tab:top_five_and_bottom_five_types}
    \begin{subtable}{\columnwidth}
        \begin{tabu}{X[1.6, l] X[1, r] X[1, r] X[1, r] X[1, r]}
            \toprule
            \textbf{Type} & \textbf{F$_1$ Score} & \textbf{Precision} & \textbf{Recall} & \textbf{Support} \\ \midrule
            \multicolumn{5}{c}{\textit{Top 5 Types}}                                                             \\ \midrule
            \semantictype{Grades}        & 0.991             & 0.989              & 0.994           & 1765             \\
            \semantictype{ISBN}          & 0.986             & 0.981              & 0.992           & 1430             \\
            \semantictype{Birth Date}    & 0.970             & 0.965              & 0.975           & 479              \\
            \semantictype{Industry}      & 0.968             & 0.947              & 0.989           & 2958             \\
           \semantictype{Affiliation}   & 0.961             & 0.966              & 0.956           & 1768             \\ \midrule
            \multicolumn{5}{c}{\textit{Bottom 5 Types}}                                                          \\ \midrule
            \semantictype{Brand}         & 0.685             & 0.760              & 0.623           & 574              \\
            \semantictype{Person}        & 0.630             & 0.654              & 0.608           & 579              \\
            \semantictype{Director}      & 0.537             & 0.700              & 0.436           & 225              \\
            \semantictype{Sales}         & 0.514             & 0.568              & 0.469           & 322              \\
            \semantictype{Ranking}       & 0.468             & 0.612              & 0.349           & 439              \\ \bottomrule
        \end{tabu}
    \end{subtable}
\end{table}

\begin{table}[h]
\caption{Examples of low precision and low recall types.}
\label{tab:examples_error}
    \begin{subtable}{\columnwidth}
        \label{tab:examples_incorrect_precision_recall}    
        \begin{tabu}{Xll}
            \toprule
            \textbf{Examples} & \textbf{True type} & \textbf{Predicted type} \\ \midrule  \addlinespace[0.1em]
            \multicolumn{3}{c}{\textit{Low Precision}} \\ \addlinespace[0.1em] \hline
            81, 13, 3, 1 & \texttt{Rank} & \texttt{Sales} \\
            316, 481, 426, 1, 223 & \texttt{Plays} & \texttt{Sales} \\
            \$, \$\$, \$\$\$, \$\$\$\$, \$\$\$\$\$ & \texttt{Symbol} & \texttt{Sales} \\ \hline
            \addlinespace[0.1em]
            \multicolumn{3}{c}{\textit{Low Recall}} \\ \addlinespace[0.1em] \hline
            \#1, \#2, \#3, \#4, \#5, \#6 & \texttt{Ranking} & \texttt{Rank} \\
            3, 6, 21, 34, 29, 36, 54 & \texttt{Ranking} & \texttt{Plays} \\
            1st, 2nd, 3rd, 4th, 5th & \texttt{Ranking} & \texttt{Position} \\ \bottomrule
        \end{tabu}
    \end{subtable}
\end{table}

To understand types for which Sherlock performs poorly, we include incorrectly predicted examples for the lowest precision type (\semantictype{sales}) and the lowest recall type (\semantictype{ranking}) in Table~\ref{tab:examples_error}. From the three examples incorrectly predicted as \semantictype{sales}, we observe that purely numerical values or values appearing in multiple classes (\eg currency symbols) present a challenge to type detection systems. From the three examples of incorrectly predicted \semantictype{ranking} columns, we again note the ambiguity of numerical values.

\subsection{Contribution by Feature Category} 

We trained feature-specific subnetworks in isolation and report the F$_1$ scores in Table~\ref{tab:scores_featuresets}. Word embedding, character distribution, and paragraph vector feature sets demonstrate roughly equal performance to each other, and significantly above that of the global statistics features, though this may be due to fewer features. Each feature set in isolation performs significantly worse than the full model, supporting our combining of each feature set.

\begin{table}[h]
    \caption{Performance contribution of isolated feature sets.}
    \label{tab:scores_featuresets}
    \begin{tabu}{X r r}
    \toprule
        \textbf{Feature set} & \textbf{Num. Features} & \textbf{F$_1$ Score} \\ \hline
        Word embeddings & 201 & $0.79$ \\
        Character distributions & 960 & $0.78$ \\
        Paragraph vectors & 400 & $0.73$ \\
        Global statistics & 27 & $0.25$ \\  \bottomrule
    
    \end{tabu}
\end{table}

\subsection{Feature Importances} 

We measure feature importance by the total reduction of the Gini impurity criterion brought by that feature to the decision tree model. The top 10 most important features from the global statistics and character-level distributions sets are shown in Table~\ref{tab:dt_importances}. While word embedding and paragraph vector features are important, they are difficult to interpret and are therefore omitted.

\begin{table}[h]
\caption{Top-10 features for the decision tree model. ``Score'' denotes normalized gini impurity.}
\label{tab:dt_importances}  
    \begin{subtable}{\columnwidth}
        \caption{Top-10 global statistics features (out of 27).}
        \label{tab:dt_importances_global}    
        \begin{tabu}{r X r}
            \toprule
            \textbf{Rank} & \textbf{Feature Name} & \textbf{Score} \\ \midrule
            1    & Number of Values             & 1.00     \\
            2    & Maximum Value Length             & 0.79  \\
            3    & Mean \# Alphabetic Characters in Cells & 0.43  \\
            4    & Fraction of Cells with Numeric Characters     & 0.38  \\
            5    & Column Entropy               & 0.35  \\
            6    & Fraction of Cells with Alphabetical Characters      & 0.33  \\
            7    & Number of None Values        & 0.33  \\
            8    & Mean Length of Values           & 0.28  \\
            9    & Proportion of Unique Values  & 0.22  \\
            10   & Mean \# of Numeric Characters in Cells    & 0.16  \\ \bottomrule
        \end{tabu}
    \end{subtable}
    
    \vspace{0.3cm}

    \begin{subtable}{\columnwidth}
        \caption{Top-10 character-level distribution features (out of 960).}
        \label{tab:dt_importances_character}    
        \begin{tabu}{r X r}
            \toprule
            \textbf{Rank} & \textbf{Feature Name}       & \textbf{Score} \\ \midrule
            1                & Sum of `D' across values    & 1.00              \\
            2                & Mean number of `M'          & 0.77           \\
            3                & Minimum number of `-'       & 0.69           \\
            4                & Skewness of `,'             & 0.59           \\
            5                & Whether all values have a `,'  & 0.47           \\
            6                & Maximum number of `g'       & 0.45           \\
            7                & Skewness of `{]}'           & 0.45           \\
            8                & Mean number of `,'          & 0.40           \\
            9                & Mean number of `z'          & 0.37           \\
            10               & Sum of `n'                  & 0.36           \\ \bottomrule
        \end{tabu}
    \end{subtable}
\end{table}

Inspecting Table~\ref{tab:dt_importances_global}, we find that the ``number of values'' in a column is the most important feature. Certain classes like \semantictype{name} and \semantictype{requirements} tended to contain fewer values, while others like \semantictype{year} and \semantictype{family} contained significantly more values. The second most important feature is the ``maximum value length'' in characters, which may differentiate classes with long values, such as \semantictype{address} and \semantictype{description}, from classes with short values, such as \semantictype{gender} and \semantictype{year}.

The top character-level distribution features in Table~\ref{tab:dt_importances_character} suggest the importance of specific characters for differentiating between types. The third most important feature, the ``minimum number of `-' characters'', likely helps determine datetime-related types. The fifth most important feature, ``whether all values have a `,' character'' may also distinguish datetime-related or name-related types. Further study of feature importances for semantic type detection is a promising direction for future research.

\subsection{Rejection Curves}
\begin{figure}[h]
    \centering
    \hspace*{-0.1cm}
    \includegraphics[width=\columnwidth]{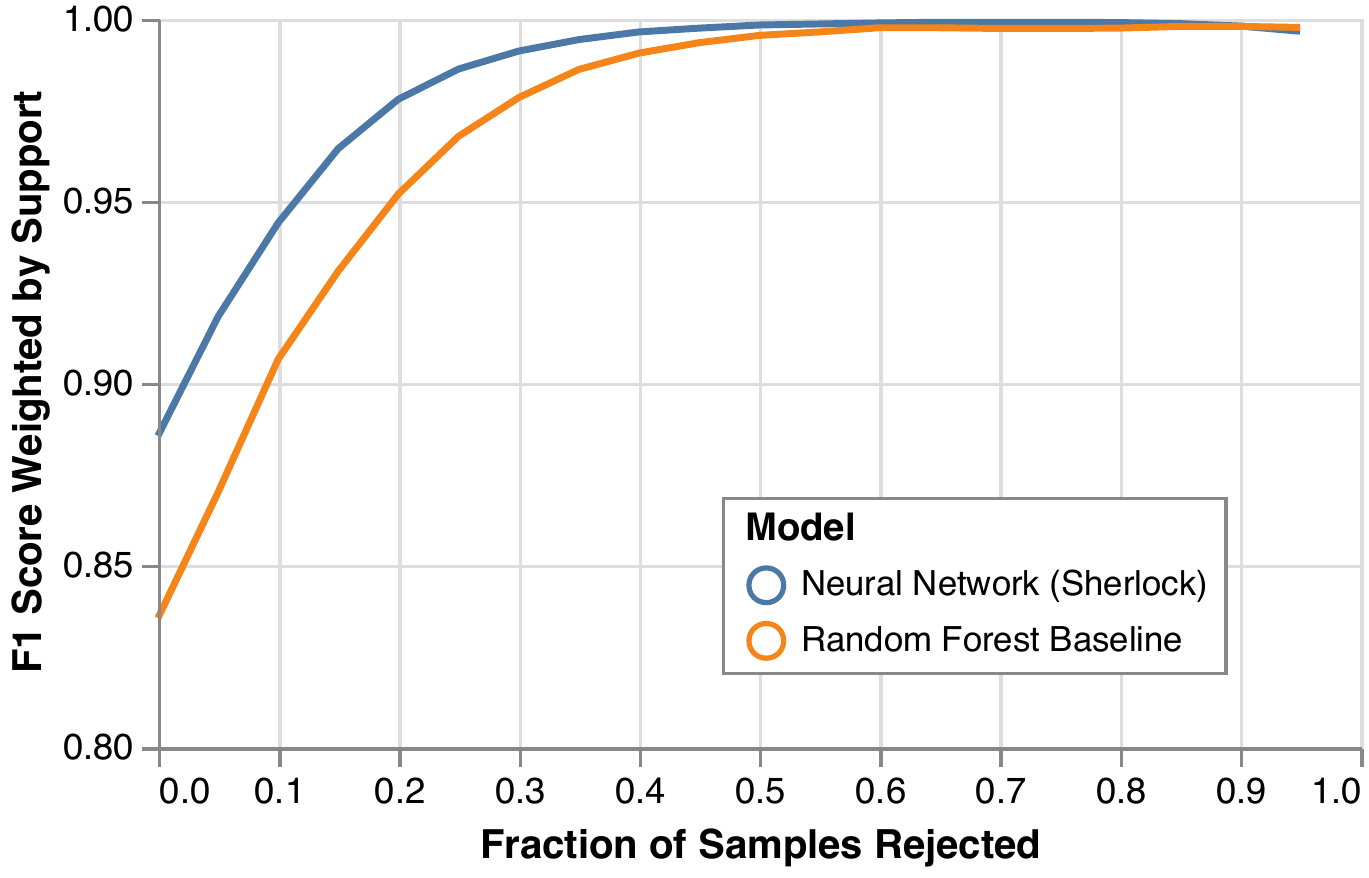}
    \caption{Rejection curves showing performance while rejecting all but the top $x\%$ highest confidence samples.}
    \label{fig:reject_curve}
    \vspace*{-0.1cm}
\end{figure}

Given unseen data values, Sherlock assesses the probability of those values belonging to each type, then predicts the type with the highest probability. Interpreting probabilities as a measure of confidence, we may want to only label samples with high confidence of belonging to a type. To understand the effect of confidence threshold on predictive performance, we present the error-rejection curves of Sherlock and the decision tree model in Figure~\ref{fig:reject_curve}.

By introducing a rejection threshold of $10$\% of the samples, Sherlock reaches an F$_1$ score of ${\sim}0.95$. This significant increase in predictive performance suggests a hybrid approach in which low confidence samples are manually annotated. Note that the higher rejection threshold, the lower the error we make in predicting labels, at the cost of needing more expert capacity.

%% file: 6_discussion.tex
\section{Discussion}

We began by considering a set of semantic types described by prior work that identifies correspondences between DBPedia~\cite{dbpedia} and WebTables~\cite{webtables}. Then, we constructed a dataset consisting of matches between those types with columns in the VizNet~\cite{hu2019viznet} corpus. Inspection of these columns suggests that such an approach yields training samples with few false positives. After extracting four categories of features describing the values of each column, we formulate type detection as a multiclass classification task.

A multi-input neural network demonstrates high predictive performance at the classification task compared to machine learning, matching-based, and crowdsourced benchmarks. We note that using real-world data provides the examples needed to train models that detect many types, at scale. We also observe that the test examples frequently include dirty (\eg missing or malformed) values, which suggests that real-world data also affords a degree of robustness. 
Measuring and operationalizing these two benefits, especially with out-of-distribution examples, is a promising direction of research.

Developers have multiple avenues to incorporating ML-based semantic type detection approaches into systems. To support the use of Sherlock ``out-of-the-box,'' we distribute Sherlock as a Python library\footnotemark[3] that can be easily installed and incorporated into existing codebases. For developers interested in a different set of semantic types, we open source our training and analysis scripts.\footnotemark[2] The repository also supports developers wishing to retrain Sherlock using data from their specific data ecologies, such as enterprise or research settings with domain-specific data.

To close, we identify four promising avenues for future research: (1) enhancing the quantity and quality of the training data, (2) increasing the number of considered types, (3) enriching the set of features extracted from each column, 
and (4) developing shared benchmarks.

\paragraph{Enhancing data quantity and quality} 
Machine learning model performance is limited by the number of training examples. Sherlock is no exception. Though the VizNet corpus aggregates datasets from four sources, there is an opportunity to incorporate training examples from additional sources, such as Kaggle,\footnote{\url{https://www.kaggle.com/datasets}} datasets included alongside the R statistical environment,\footnote{\url{https://github.com/vincentarelbundock/Rdatasets}} and the ClueWeb web crawl of Excel spreadsheets.\footnote{\url{http://lemurproject.org/clueweb09.php}} We expect increases in training data diversity to improve the robustness and generalizability of Sherlock.

Model predictions quality is further determined by the correspondence between training data and unseen testing data, such as datasets uploaded by analysts to a system. Our method of matching semantic types with columns from real-world data repositories affords both the harvesting of training samples at scale and the ability to use aspects of dirty data, such as the number of missing values, as features. While we verified the quality of training data through manual inspection, there is an opportunity to label data quality at scale by combining crowdsourcing with active learning. By assessing the quality of each training dataset, such an approach would support training semantic type detection models with completely ``clean'' data at scale.

\paragraph{Increasing number of semantic types}
To ground our approach in prior work, this paper considered 78 semantic types described by the T2Dv2 Gold Standard. While $78$ semantic types is a substantial increase over what is supported in existing systems, it is a small subset of entities from existing knowledge bases: the DBPedia ontology~\cite{dbpedia} covers 685 classes, WordNet~\cite{wordnet} contains $175$K synonym sets, and Knowledge Graph\footnote{\url{https://developers.google.com/knowledge-graph}} contains millions of entities. The entities within these knowledge bases, and hierarchical relationships between entities, provide an abundance of semantic types.

In lieu of a relevant ontology, researchers can count frequency of column headers in available data to determine which semantic types to consider. Such a data-driven approach would ensure the maximum number of training samples for each semantic type. Additionally, these surfaced semantic types are potentially more specific to usecase and data ecology, such as data scientists integrating enterprise databases within a company.

\paragraph{Enriching feature extraction}
We incorporate four categories of features that describe different aspects of column values.
A promising approach is to include features that describe relationships between columns (\eg correlation, number of overlapping values, and name similarity), aspects of the entire dataset (\eg number of columns), and source context (\eg webpage title for scraped tables).
Additionally, while we used features to aid interpretation of results, neural networks using raw data as input are a promising direction of research. For example, a character-level recurrent neural network could classify concatenated column values.

\paragraph{Developing shared benchmarks}
Despite rich prior research in semantic type detection, we could not find a benchmark with publicly available code that accommodates a larger set of semantic types. We therefore incorporated benchmarks that approximated state-of-the-art data systems, to the best of our knowledge. However, domains such as image classification and language translation have benefited from shared benchmarks and test sets. Towards this end, we hope that open-sourcing the data and code used in this paper can benefit future research.

%% file: 8_conclusion.tex
\section{Conclusion}
Correctly detecting semantic types is critical to many important data science tasks. Machine learning models coupled with large-scale data repositories have demonstrated success across domains, and suggest a promising approach to semantic type detection. Sherlock provides a step forward towards this direction.

%% file: A_appendix.tex
\section{Appendix}

\subsection{Supplemental Tables}\label{app:classes}

\vspace{-0.2cm}
\begin{table}[h!]
    \caption{$78$ semantic types included in this study.}
    \label{tab:classes}
    \small
    \hspace*{-0.3cm}
    \begin{tabu}{lllll}
    \toprule
        \multicolumn{5}{c}{{\normalsize\textbf{Semantic Types}}} \\ \hline
        Address         &  Code         &  Education    & Notes         & Requirement\\
        Affiliate       &  Collection   &  Elevation    & Operator      & Result\\
        Affiliation     &  Command      &  Family       & Order         & Sales\\
        Age             &  Company      &  File size    & Organisation  & Service\\
        Album           &  Component    &  Format       & Origin        & Sex\\
        Area            &  Continent    &  Gender       & Owner         & Species\\
        Artist          &  Country      &  Genre        & Person        & State\\
        Birth date      &  County       &  Grades       & Plays         & Status\\
        Birth place     &  Creator      &  Industry     & Position      & Symbol\\
        Brand           &  Credit       &  ISBN         & Product       & Team\\
        Capacity        &  Currency     &  Jockey       & Publisher     & Team name\\
        Category        &  Day          &  Language     & Range         & Type\\
        City            &  Depth        &  Location     & Rank          & Weight\\
        Class           &  Description  &  Manufacturer & Ranking       & Year\\
        Classification  &  Director     &  Name         & Region        & \\
        Club            &  Duration     &  Nationality  & Religion      & \\
    \bottomrule
    \end{tabu}
\end{table}

\vspace{-0.2cm}
\begin{table}[h!]
    \centering
    \caption{Description of the $27$ global statistical features. Asterisks (*) denote features included in Venetis et al.~\cite{venetis2011recovering}.}
    \label{tab:character-features}
    \begin{tabu}{X}
    \toprule
        \textbf{Feature description} \\ \hline
        Number of values. \\
        Column entropy. \\
        Fraction of values with unique content.* \\
        Fraction of values with numerical characters.* \\
        Fraction of values with alphabetical characters. \\ 
        Mean and std. of the number of numerical characters in values.* \\
        Mean and std. of the number of alphabetical characters in values.* \\
        Mean and std. of the number special characters in values.* \\   
        Mean and std. of the number of words in values.* \\
        \{Percentage, count, only/has-Boolean\} of the None values. \\
        \{Stats, sum, min, max, median, mode, kurtosis, skewness,\\any/all-Boolean\} of length of values.\\
    \bottomrule    
    \end{tabu}
\end{table}

\subsection{Mechanical Turk Demographics}\label{app:mturk_demographics}
Of the 390 participants, 57.18\% were male and 0.43\% female. 1.5\% completed some high school without attaining a diploma, while others had associates (10.5\%), bachelor's (61.0\%), master's (13.1\%), or doctorate or professional degree (1.8\%) in addition to a high school diploma (12.3\%). 26.4\% of participants worked with data daily, 33.1\% weekly, 17.2\% monthly, and 11.0\% annually, while 12.3\% never work with data. In terms of age: 10.0\% of participants were between 18-23,  24-34 (60.3\%), 35-40 (13.3\%), 41-54 (12.6\%), and above 55 (3.8\%).